\documentclass[twoside,11pt]{article}

\usepackage{melba}

\usepackage{adjustbox}
\usepackage[flushleft]{threeparttable} 
\usepackage{booktabs,caption}
\usepackage[utf8]{inputenc}
\usepackage[english]{babel}
\usepackage{fancyhdr}
\usepackage{lastpage}
\usepackage{mwe} 

\usepackage{amssymb}
\usepackage{latexsym}
\usepackage{subfig}
\usepackage{array}

\usepackage{lscape} 
\usepackage{siunitx,tabularx,ragged2e,booktabs,caption}
\usepackage{adjustbox}
\usepackage{framed,multirow}

\if@abbrvbib
\fi

\usepackage[colorinlistoftodos,prependcaption,textsize=small]{todonotes}

\usepackage{amsmath,amsfonts}

\melbaid{YYYY:NNN}  
\doi{https://doi.org/10.59275/j.melba.2023-AAAA}
\melbaauthors{Dawood et al}  
\volume{2}
\firstpageno{1337}  
\melbayear{YYYY}  
\datesubmitted{m1/yyyy}  
\datepublished{m2/yyyy}  

\melbaspecialissue{Medical Imaging with Deep Learning (MIDL) 2020}
\melbaspecialissueeditors{Marleen de Bruijne, Tal Arbel, Ismail Ben Ayed, Hervé Lombaert}

\ShortHeadings{Improving Deep Learning Model Calibration}{Dawood et al}

\title{Improving Deep Learning Model Calibration for Cardiac Applications using Deterministic Uncertainty Networks and Uncertainty-aware Training}

\author{\firstname Tareen \surname Dawood \orcid{0000-0003-4932-8659} \email tareen.dawood@kcl.ac.uk \\  
	\addr School of Biomedical Engineering \& Imaging Sciences, King\textquotesingle s College London, UK
	\AND
	\name Bram Ruijsink\orcid{0000-0001-8313-5709} \email jacobus.ruijsink@kcl.ac.uk \\
	\addr School of Biomedical Engineering \& Imaging Sciences, King\textquotesingle s College London, UK
        \addr Guy\textquotesingle{}s and St Thomas\textquotesingle{} Hospital, London, UK
 	\AND
	\name Reza Razavi\orcid{0000-0003-1065-3008} \email reza.razavi@kcl.ac.uk\\
	\addr School of Biomedical Engineering \& Imaging Sciences, King\textquotesingle s College London, UK
 	\AND
	\name Andrew P. King*\orcid{0000-0002-9965-7015} \email andrew.king@kcl.ac.uk\\
	\addr School of Biomedical Engineering \& Imaging Sciences, King\textquotesingle s College London, UK
  	\AND
	\name Esther Puyol-Ant\'on*\orcid{0000-0002-9789-6629} \email esther.puyol\_anton@kcl.ac.uk\\
	\addr School of Biomedical Engineering \& Imaging Sciences, King\textquotesingle s College London, UK\\
* Joint last authors
}

\begin{document}
\maketitle
\begin{abstract}

Improving calibration performance in deep learning (DL) classification models is important when planning the use of DL in a decision-support setting. In such a scenario, a confident wrong prediction could lead to a lack of trust and/or harm in a high-risk application. We evaluate the impact on accuracy and calibration of two types of approach that aim to improve DL classification model calibration:  deterministic uncertainty methods (DUM) and uncertainty-aware training. Specifically, we test the performance of three DUMs and two uncertainty-aware training approaches as well as their combinations. To evaluate their utility, we use two realistic clinical applications from the field of cardiac imaging: artefact detection from phase contrast cardiac magnetic resonance (CMR) and disease diagnosis from the public ACDC CMR dataset. Our results indicate that both DUMs and uncertainty-aware training can improve both accuracy and calibration in both of our applications, with DUMs generally offering the best improvements. We also investigate the combination of the two approaches, resulting in a novel deterministic uncertainty-aware training approach. This provides further improvements for some combinations of DUMs and uncertainty-aware training approaches.
\end{abstract}

\section{Introduction}
\label{sec:Introduction}
 
Deep learning (DL) classification models have been found to be poorly \emph{calibrated} with respect to the uncertainty of their predictions \citep{guo2017calibration}. In other words, their certainty does not always match their accuracy, leading to, for example, certain but wrong model predictions.  This problem has been attributed to high-capacity DL models over-fitting to conventional loss functions \citep{mukhoti2020calibrating} or feature collapse \citep{mukhoti2021deterministic}. Whatever the underlying cause, when considering the use of DL in a decision support setting, this creates obvious concerns about the model providing misleading information to the decision maker, thus negatively affecting trust and potentially causing harm in high-risk settings such as healthcare \citep{ding2020revisiting}. Standard performance metrics such as accuracy can create an impression of a well-performing model, even when calibration is poor. Therefore, for a more comprehensive understanding of model performance additional metrics need to be reported \citep{maier2206metrics}.

A number of techniques have been proposed to address the issue of poor calibration. These mostly focus on one of three types of approach: (i) post-hoc calibration \citep{guo2017calibration}, which involves modifying the probabilistic outputs of a poorly calibrated model, (ii) uncertainty-aware training \citep{krishnan2020improving, kumar2018trainable, karandikar2021soft, mukhoti2021deep,dawood2023uncertainty}, in which the training of the model is altered (e.g. via the loss function) to improve calibration, and (iii) improving the robustness of the uncertainty estimates. This third type of approach can encompass ensemble-based and sampling-based uncertainty estimation \citep{lakshminarayanan2017simple, thiagarajan2022training}, as well as the more recently proposed deterministic uncertainty methods (DUM) \citep{mukhoti2021deep, franchi2022latent}. Of these three types of approach, post-hoc calibration is primarily a way to fix poorly calibrated models and is known to lack robustness in the presence of domain shift \citep{wang2023calibration}. Sampling-based and ensemble-based methods can improve calibration but have computational costs and are also a way to get better uncertainty estimates from poorly trained models \citep{zou2023review, buddenkotte2023calibrating}. Therefore, our focus in this paper will be on methods to improve the training of DL models, namely uncertainty-aware training and DUMs, which we believe represent the most promising ways forward for achieving trustworthy and well-calibrated classification models.

To date, there has been no thorough evaluation of the comparative effectiveness of uncertainty-aware training and DUMs in improving model calibration on real medical imaging problems. The first aim of this paper is to perform such an evaluation on two cardiac magnetic resonance (CMR) image classification problems. Furthermore, with the exception of our preliminary work \citep{dawood2023addressing}, there has been no investigation of the effectiveness of combining uncertainty-aware training with DUMs in the medical imaging domain. This is the second aim of this paper.


\section{Related Works}

\subsection{Model Calibration}
\label{calibration_section}

Calibration refers to how faithfully a model's predicted classification probability can estimate the true correctness and it can be measured by examining the relationship between a model's certainty and its accuracy \citep{guo2017calibration, wang2023calibration}. Various metrics have been investigated to measure calibration, and several advantages and pitfalls of these have been identified in research by \cite{maier2206metrics} and in our previous work \citep{dawood2023uncertainty}. 

Expected Calibration Error (ECE) is the most widely used metric to assess calibration, although recent work has shown it to be a biased empirical measure of calibration \citep{roelofs2022mitigating}. Thus, evaluating calibration using this metric alone may not provide a complete picture of model calibration performance. In \cite{dawood2023uncertainty}, we used a diverse set of calibration metrics (ECE, Adaptive ECE (AECE), Overconfidence Error (OE), Maximum Calibration Error (MCE), and Brier Score (BS)) and found that they often provided different conclusions regarding model calibration. Interestingly, the findings suggested that the OE metric, which only measures when confidence exceeds accuracy, may be more applicable for high-risk healthcare settings, aligning with previous work \citep{thulasidasan2019mixup}. AECE aims to address biases introduced by the hard binning strategy used by ECE, and instead adapts its bin sizes based on the distribution of probabilities in the sample. Therefore, although less widely used than ECE, AECE may be a preferable metric in many ways. However, investigating different calibration metrics is still an ongoing research area \citep{maier2206metrics}, and it may be that there is no single calibration metric that will be robust and applicable to all situations.

\subsection{Improving Calibration}

\subsubsection{Post-hoc Methods}

\emph{Post-hoc calibration} involves scaling the predicted classification probabilities (which are considered to be the complement of the uncertainty) to recalibrate the model outputs. A well-known post-hoc method is Platt scaling \citep{platt1999probabilistic} which transforms the logits of a trained model using a logistic regression model. An alternative to Platt scaling is Temperature Scaling (TS), which scales the logits using a temperature parameter before feeding them into a Softmax function \citep{guo2017calibration}. Interestingly, TS seems to be favoured in a data-limited regime \citep{zhang2020mix}, whilst other research suggests it may only be effective if used alongside Deep Ensembles \citep{rahaman2021uncertainty}. In addition, as stated earlier, post-hoc methods are known to lack robustness to domain shift \citep{wang2023calibration}.

\subsubsection{Uncertainty Aware Training}
\label{UCA}

Uncertainty-aware training of DL models involves incorporating estimates of uncertainty into the model training, with the aim of improving calibration. To date, this has mostly involved using the Softmax class probability as the model confidence (and its complement as the uncertainty). For example, in \cite{dawood2023uncertainty} we evaluated various novel and state-of-the-art uncertainty-aware training strategies for cardiac imaging applications. Examples of state-of-the-art approaches to uncertainty-aware training include Accuracy versus Uncertainty (AvUC) \citep{krishnan2020improving}, and Maximum Mean Calibration Error (MMCE) \citep{kumar2019calibration}, which are both differentiable loss terms aimed at improving model calibration. AvUC works by placing every prediction into one of four categories: accurate and certain, accurate and uncertain, inaccurate and certain, and inaccurate and uncertain. The MMCE loss utilises a reproducing kernel Hilbert space (RKHS) to create a theoretically sound measure of calibration for use in model training to improve calibration performance. An alternative uncertainty-aware training approach \citep{yu2019uncertainty} utilised a student-teacher framework, in which the student model gradually learned to produce calibrated predictions by exploiting uncertainty information for a 3D left atrium segmentation application. In other work, \cite{9214845} utilised an ensembling strategy alongside an uncertainty-aware unsupervised loss to improve the reliability of breast mass segmentation from ultrasound. These works indicate the need to incorporate uncertainty information into DL model training to improve calibration, which may help develop more reliable DL models for healthcare applications \citep{gawlikowski2021survey}. 

\subsubsection{Improving Uncertainty Estimates}

\subsubsection*{Non-deterministic Methods}

Bayesian neural networks learn a probability distribution over network weights and allow estimation of uncertainty through variational inference or sampling-based approaches like Monte Carlo dropout. Alternatively, ensemble approaches are an approximation to a full Bayesian neural network and involve training multiple models whose outputs can be combined to estimate uncertainty. These approaches are referred to as non-deterministic because they require multiple strategies to produce multiple model outputs, which are then used to quantify the model (un)certainty \citep{abdar2021review}. Recent work by \cite{buddenkotte2023calibrating} et al. on medical image segmentation found that ensembles of identical models trained with different random seeds failed to approximate the classification probability. Their proposed solution was to train an ensemble of models ranging from highly sensitive to highly precise, enabling more robust and calibrated estimates of probability. However, ensemble-based approaches like this can be computationally expensive due to the need to train multiple models \citep{abdar2021review, gawlikowski2021survey, wang2023calibration}.

\subsubsection*{Deterministic Methods}

Compared to Bayesian neural networks and ensemble-based approaches, DUMs train only a single DL model and perform a single pass to generate the model output and an uncertainty estimate. The computational cost of such models is thus greatly reduced. Two approaches can be used to quantify uncertainty in DUMs: internal, using a single network to predict the uncertainty, or external, using additional components to estimate the uncertainty \citep{gawlikowski2021survey}. Both of these methods have the potential to differentiate between different types of uncertainty using a single forward pass by varying the network architecture to capture both aleatoric and epistemic uncertainty \citep{gawlikowski2021survey}.

Examples of DUMs include Evidential Neural Networks (ENN), Prior Networks, Deep Deterministic Uncertainty (DDU) networks, and Latent Discriminant Deterministic Uncertainty (LDU) networks \citep{abdar2021review, mukhoti2021deep, franchi2022latent}. These networks offer promising approaches aimed at reducing the associated weaknesses of feature collapse even when evaluating a standard neural network with the Softmax function. 

\paragraph{Dirichlet Distributions and Evidential Neural Networks:}

The Dirichlet distribution models class probabilities as points on a simplex, in which each point is a categorical distribution over class labels conditioned on the input data \citep{malinin2018predictive}. 
Dirichlet distributions have been used to develop ENNs using subjective logic and evidential theory \citep{sensoy2018evidential,malinin2018predictive,sensoy2021misclassification}. ENN models provide a probability or uncertainty estimate per class and optimise the parameters of the network by introducing a loss formula dependent on the evidence and final beliefs allocated to each class/category.

\paragraph{Deep Deterministic Uncertainty Networks:}

DDU networks are deterministic networks with minimal architectural changes that have been shown to both perform as well as a Deep Ensemble for uncertainty prediction and disentangle uncertainty when evaluated on out-of-distribution (OOD) data. Building on previous work in Generative Adversarial Networks (GAN's), \cite{mukhoti2021deep} have shown that spectral normalisation layers placed on the fully connected layers of a neural network work can alleviate feature collapse in the latent space by encouraging smoothness and sensitivity \citep{mukhoti2021deep}. Interestingly, regularising the latent space with these additional architectural changes has shown improvements in calibration, even with the Softmax function, but has not been extensively evaluated in different application domains \citep{postels2021practicality}.

\paragraph{Latent Discriminant Deterministic Uncertainty Networks:}

LDU networks aim to enhance the deterministic capabilities found in DDUs, but use a distinction maximisation layer placed over the latent space to obtain a representative class of prototypes \citep{franchi2022latent}. The prototypes help to separate input samples by acting as knowledge or references. The network's architectural design, when compared to ENNs and DDUs, is unique, as typically the DUM approach works by replacing the last layers of the network, whereas LDU utilises the distinction maximisation as a hidden layer (using prototypes) over the latent space, to both guide and preserve the discriminative properties of the latent representations. In this way, the model can optimise its final classification decision, as differentiation between classes should be more distinct \citep{franchi2022latent}. 

\subsection{Contributions}

In this paper, we perform an investigation into the best way to improve model calibration during training using two real-world clinical applications from medical imaging. Our specific contributions are:

\begin{enumerate}
  \item We evaluate three different DUM approaches and two different uncertainty-aware training methods, and also investigate the novel combination of DUMs with uncertainty-aware training.
  \item All methods are evaluated on two realistic medical imaging applications: artefact detection and disease diagnosis, both from CMR images.
  \item We use calibration performance measures proposed in the literature and previous findings to understand the effects of different approaches on calibration.
\end{enumerate}

This paper is structured as follows. In Section \ref{sec:Method} we describe the baseline and DUM models, as well as the uncertainty-aware training strategies. In Section \ref{sec:Material_Exp} we present all experiments performed to evaluate and compare the different approaches and present all results in Section \ref{sec:Results}. Section \ref{sec:Discussion} discusses the findings, evaluates the outcomes, and recommends future work towards cultivating trustworthy and calibrated DL classification models.

\section{Methods}
\label{sec:Method}
In this section, we provide details of our baseline model and describe the adaptations that we made to it to implement the three DUMs and two uncertainty-aware training strategies evaluated in this paper. We also describe how the DUMs can be combined with uncertainty-aware training.

\subsection{Baseline Model}
\label{sec:baseline}
\begin{figure*}[ht]
    \includegraphics[width=\textwidth]{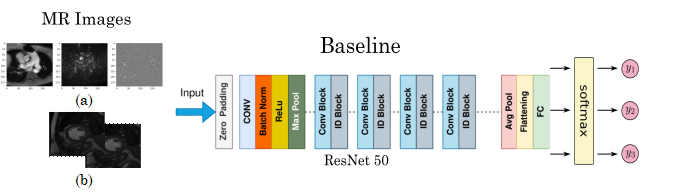}
    \centering
    \caption{Illustration of the baseline classification model with a 3D ResNet50 architecture. Images in (a) represent the phase contrast cardiac magnetic resonance (CMR) data and (b) the ACDC CMR dataset respectively.}
    \label{fig:base_model}%
\end{figure*}

The 3D ResNet50 architecture was used as the baseline classification model, as illustrated in Figure \ref{fig:base_model}.
The ResNet50 was trained using the standard cross entropy loss ($\mathcal{L}_{\mathrm{CE}}$) between the predicted probabilities and ground truth labels. The cross-entropy loss function is defined as:

\begin{equation} 
\centering
\large
    \mathcal{L}_{\text{CE}}
    = -\sum_{c=1}^M y_{o,c}\log(p_{o,c})
    \label{eqn:ce_loss}  
\end{equation}
where $\mathrm{\emph{M}}$ is the number of classes, $\mathrm{\emph{y}}$ the class label in the classification model, and $\mathrm{\emph{p}}$ the predicted probability for sample $\mathrm{\emph{o}}$.

\subsection{Deterministic Uncertainty Models}

We implemented and evaluated three state-of-the-art DUMs and details of these are provided in the sections below.

\subsubsection{Evidential Neural Network}
\label{sec:ENN}

The ENN \citep{sensoy2018evidential} was implemented by replacing the Softmax layer of the baseline model described in Section \ref{sec:baseline} with a ReLU activation function, the outputs of which were used as parameters (evidence) for the Dirichlet distribution as illustrated in Figure \ref{fig:All_Architecture}a. Additionally, the cross entropy loss function was replaced to allow the model to quantify the expected risk associated with choosing a particular class/category during training and inference. The ENN loss function was defined as described in \cite{sensoy2018evidential, sensoy2020uncertainty}:
\begin{equation} 
\centering
\large
\mathcal{L_{\text{i}}}(\mathrm{\theta}) = \sum_{c=1}^My_{o,c}\bigg(\psi(\text{S}_i) - \psi(\alpha_{ij})\bigg) 
    \label{eqn:enn_loss}  
\end{equation}
where, as in the baseline model notation, $M$ is the number of classes and $o$ is the sample. In addition, $\mathrm{\alpha}$ represents the parameters of the Dirichlet distribution, $\mathrm{S}$ the Dirichlet strength, and $\mathrm{\psi}$ the digamma function. 

An additional term was then added to the loss function, $\mathcal{D}_{\text{KL}}$, to regularise the predictive distribution by penalising the model when it diverges from a ``I do not know'' state \citep{sensoy2018evidential}. A hyperparameter {\textbf{$\lambda_{ENN}$}} is utilised to determine the influence of this term. Further details and derivations can be found in the original paper by \cite{sensoy2018evidential}.

\begin{equation} 
\centering
\large
    \mathcal{L}_{\text{ENN}} = 
    \mathcal{L_{\text{i}}}(\mathrm{\theta}) + \textbf{$\lambda_{ENN}$}\mathcal{D}_{\text{KL}}
    \label{eqn:enn_total_loss}  
\end{equation}

\subsubsection{Deep Deterministic Uncertainty}
\label{sec:DDU}

In this approach, the same loss is applied as in Section \ref{sec:baseline} but the structure of the model is changed, with spectral normalisation added into the fully connected layers as illustrated in Figure \ref{fig:All_Architecture}b. Spectral normalisation was first proposed by \cite{miyato2018spectral} and tested on simple image datasets such as CIFAR10. It is a weight normalisation that constrains the spectral norm of each layer in a network and in this way has been proven to regularise the feature space well and improve discrimination between classes \citep{mukhoti2021deep}. To achieve this the DDU has two tunable hyperparameters, the spectral normalisation coefficient, {\textbf{$\mathrm{SN_{coeff}}$}}, and the stochastic gradient descent momentum, {\textbf{$\mathrm{S_{m}}$}}, as in \cite{mukhoti2021deep}.

\subsubsection{Latent Discriminant Deterministic Uncertainty}
\label{sec:LDU}

\begin{figure*}[ht]
    \includegraphics[width=\textwidth]{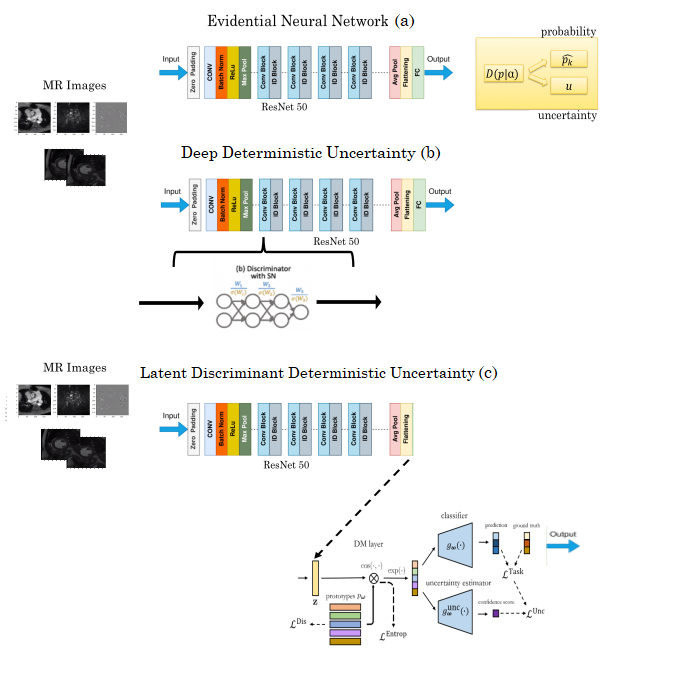}
    \centering
    \caption{Adaptations to the baseline classification model (Figure \ref{fig:base_model}) to implement the three deterministic uncertainty methods. (a) Evidential neural network, (b) Deep deterministic uncertainty network, (c) Latent discriminant deterministic uncertainty network. See text for further details.}
    \label{fig:All_Architecture}%
\end{figure*}

In LDU networks multiple new losses are added to the baseline model loss function, and furthermore, a distinction maximisation (DM) layer is included in the model as illustrated in Figure \ref{fig:All_Architecture}c. The DM layer replaces the standard last layer of a network to improve uncertainty estimates. \cite{franchi2022latent} explain that in the DM layer ``the units of the classification layer are seen as representative class prototypes and the classification prediction is computed by analyzing the localization of the input sample w.r.t. all class prototypes as indicated by the negative Euclidean distance. Considering this change the overall loss function is defined as:

\begin{equation} 
\centering
\large
    \mathcal{L}_{\text{LDU}}
    = \mathcal{L}_{\text{CE}}
    + \lambda_E \mathcal{L}_{\text{Entropy}}
    + \lambda_D \mathcal{L}_{\text{Dis}} 
    + \lambda_u \mathcal{L}_{\text{Unc}}
    \label{eqn:ldu_loss}  
\end{equation}
where $\mathcal{L}_{\mathrm{CE}}$ represents the standard cross-entropy loss as in the baseline model, $\mathcal{L}_{\mathrm{Entropy}}$ acts to constrain the latent representation using the new DM layer, $\mathcal{L}_{\mathrm{Dis}}$ relates to the loss with respect to the prototypes and $\mathcal{L}_{\mathrm{Unc}}$ is the loss associated with the error of the network relating prototypes and uncertainty. Thus, this model has three additional hyperparameters: $\lambda_E$, $\lambda_D$ and $\lambda_u$. Further details can be found in  \cite{franchi2022latent}. 

\subsection{Uncertainty Aware Training Strategies}
\label{sec:Comparat_Section}

We now present two existing methods from the literature as our uncertainty-aware training strategies and explain how the baseline model was adapted to implement these. 

\subsubsection{Accuracy Versus Uncertainty Loss}
\label{sbsc:AvUCLoss}

The AvUC loss \citep{krishnan2020improving} utilises the relationship between accuracy and uncertainty to develop a loss function aimed at improving model calibration. The loss function is differentiable and works by placing each prediction into one of four categories: accurate and certain (AC), accurate and uncertain (AU), inaccurate and certain (IC), and inaccurate and uncertain (IU). Utilising these four categories, the loss term is defined as follows:
\begin{eqnarray}\centering
    \mathcal{L}_{\mathrm{AvUC}}
    = & log \left( { 1 + \cfrac{n_{AU} + n_{IC}}{n_{AC} + n_{IU}}}\right)
    \label{eqn:uncertaintyloss}
    \end{eqnarray}
where $n_{AU}, n_{IC}, n_{AC}, n_{IU}$ represent the number of accurate/uncertain, inaccurate/certain, accurate/uncertain, and inaccurate/uncertain samples within each batch. For this model, the baseline model is left intact but the AvUC loss is added to the total loss, weighted by a hyperparameter $\lambda_A$.
\begin{equation} 
\centering
\large
    \mathcal{L}_{\mathrm{UCA}}
    = \mathcal{L}_{\text{CE}}
    + \lambda_A\mathcal{L}_{\mathrm{AvUC}} 
    \label{eqn:all_AvUC}  
\end{equation}
where $\mathcal{L}_{\text{CE}}$ is the loss presented in the baseline model in Section \ref{sec:baseline}.

\subsubsection{Maximum Mean Calibration Error Loss Function}
\label{sbsc:MMCE}

\cite{kumar2018trainable} utilised a RKHS approach with a differentiable loss function to improve calibration, which they termed the MMCE. The MMCE loss term is defined as:
\begin{align}
    \mathcal{L_{\text{MMCE}}}(D) = 
    \sum_{c_{i}=c_{j}=0}\frac{r_{i}r_{j}k(r_{i}r_{j})}{(m-n)(m-n)} \nonumber \\
      + \sum_{c_{i}=c_{j}=1}\frac{(1-r_{i})(1-r_{j})k(r_{i}r_{j})}{{n}^2} \nonumber \\ 
     -2 \sum_{c_{i}=1, c_{j}=0}\frac{(1-r_{i})r_{j}k(r_{i}r_{j})}{(m-n)n}
\end{align}
Here, the terms $c_i$ and $c_j$ are the correctness scores per instance, $r_i$, $r_j$ are the confidence scores and $D$ is the data. The term $m$ denotes the number of correct examples in a batch of size $n$, with $k$ representing the Hilbert space kernel. Further detailed derivations and explanations are provided in \cite{kumar2018trainable}. This loss term was combined with the baseline loss function using a hyperparameter {\textbf{$\lambda_{M}$}}:

\begin{equation} 
\centering
\large
    \mathcal{L}_{\mathrm{UCA}}
    = \mathcal{L}_{\text{CE}}
    + \textbf{$\lambda_{M}$}\mathcal{L_{\text{MMCE}}}
    \label{eqn:all_MMCE}  
\end{equation}
where again $\mathcal{L}_{\text{CE}}$ is the loss presented in the baseline model in Section \ref{sec:baseline}.

\subsection{Combining DUMs with Uncertainty-aware Training}
\label{sbsc:ExpOptim}

Finally, we present details of how the DUMs and uncertainty-aware training strategies were combined to investigate if their combination could lead to further improvements in calibration performance.

\subsubsection{Evidential Neural Network}

When combining the ENN with uncertainty-aware training the uncertainty output from the ENN was used instead of the complement of the Softmax class probabilities. Therefore, these ENN uncertainty estimates could be fed directly into the additional AvUC and MMCE loss terms.

\subsubsection{Deep Deterministic Uncertainty Network}

In the original work \citep{mukhoti2021deep}, the DDU network generated confidence values as an output and the uncertainty was estimated using the AUROC metric computed from out-of-distribution (OOD) data. For our work, in the absence of OOD data, we utilised the confidence outputs of the model in our AvUC and MMCE loss functions and where required the complement of the confidence was used as the uncertainty, i.e. in the AVUC loss term $n_{AU}$. 

\subsubsection{Latent Discriminant Deterministic Uncertainty}

We followed the work of \cite{franchi2022latent} to estimate the uncertainty values for the LDU model and used the maximum class probability to implement the uncertainty-aware training strategies as this was their approach for a classification problem. Similar to the DDU, where required, the complement of the confidence was used as the uncertainty in relevant loss terms. 

\section{Materials and Experiments}
\label{sec:Material_Exp}

Two different experiments were performed to evaluate the impact of the different methods to address the problem of model calibration. Below we describe the data used and the tasks for each experiment.

\subsection{Phase Contrast Cardiac Magnetic Resonance Artefact Detection}

For the first experiment, we utilised two retrospective 2D flow phase-contrast cardiac magnetic resonance (PC-CMR) databases, in which all images were manually annotated for the presence of artefacts by multiple cardiologists \citep{chan2022automated}. The first database contained images from 80 patients from the UK Biobank \citep{sudlow2015uk} and the second database was retrieved from the clinical imaging database of Guy’s and St Thomas’ NHS Foundation Trust (GSTFT) and contained images from 830 patients with a range of different cardiovascular diseases acquired under routine clinical practice. Further details about the data and protocol can be found in \cite{chan2022automated}. The task in this experiment was to predict the presence of image artefacts from a temporal sequence (the first 25 frames for each patient were utilised) of 2D PC-CMR images combined into a 3D stack (2D+time).

\subsection{Disease Diagnosis from Cine CMR}

For this experiment, we used the publicly available ACDC dataset consisting of 150 short-axis cine steady-state free precession (SSFP) CMR slices covering the left and right ventricles from base to apex, with each cardiac cycle covered by 28 to 40 frames \citep{bernard2018deep, wibowo2022cardiac}. In our work, we used the original images, two cardiac phases (end diastole and end systole), and the first six slices for each patient \citep{wibowo2022cardiac}. The task in this experiment was to classify between 5 different subgroups (one normal and four cardiac pathologies\footnote[1]{Previous myocardial infarction, dilated cardiomyopathy, hypertrophic cardiomyopathy, abnormal right ventricle.}) from a stack of 2D cine CMR images combined into a 3D volume. Each cardiac phase (end diastole and end systole) was considered as a separate data sample with a common diagnostic label.

\subsection{Experimental Setup and Hyperparameter Optimisation}
\label{sect:Experimental_Hyperparameter}

\subsubsection{PC-CMR} The PC-CMR databases were combined and 80 \% of the total data were used for training, 10 \% for validation, and 10 \% for testing.  The validation set was used for hyperparameter optimisation. The batch size for all PC-CMR experiments was set to 16. 

\subsubsection{ACDC} The ACDC Challenge dataset had already split the patients into 100 patients for training and 50 for testing. Therefore, as we consider end diastole and end systole as separate samples we have 200 volumes for training and 100 for testing. We performed a 3-fold cross-validation on the training set for hyperparameter optimisation, then retrained using all training data using the chosen hyperparameter values. The batch size for all ACDC experiments was set to 10.

\subsubsection {Hyperparameter Optimisation}
\label{sect:add_params}

 All hyperparameters were optimised using a grid search strategy and the best values selected to maximise validation set accuracy. This optimisation process was performed separately for each approach evaluated.
 
 The learning rate for both PC-CMR and ACDC experiments was optimised using a range of $\mathrm{10^{-3}}$ to $\mathrm{10^{-6}}$ in factors of 10 and the maximum number of epochs was optimised using a range of $\mathrm{60 - 300}$ in steps of 10.

The AvUC loss function utilised a loss weight  hyperparameter ($\lambda_A$), which was optimised in the range $\mathrm{0-5}$ in steps of 0.1. For the MMCE loss function, the weight ({\textbf{$\lambda_{M}$}, see Equation \ref{eqn:all_MMCE}}) was optimised in the range $\mathrm{0 - 50}$ in steps of 1. The ENN had a scaling factor, {\textbf{$\lambda_{ENN}$}}, which was optimised in the range $\mathrm{0-40}$ in steps of 2 and 5. For the DDU the additional hyperparameters were the spectral normalisation coefficient and the momentum, and both of these were optimised in the range $\mathrm{0-2}$ in steps of 0.1. Finally, the three loss term weightings for the LDU ({\textbf{$\mathrm{\lambda_{E}}$}}, {\textbf{$\mathrm{\lambda_{D}}$}}, {\textbf{$\mathrm{\lambda_{u}}$}}, see Equation \ref{eqn:ldu_loss}) were optimised in the range $\mathrm{0-5}$ in steps of 0.1, similar to their original implementations \citep{franchi2022latent}.

See Tables 1 and 2 for the final hyperparameter values for the PC-CMR and ACDC experiments.

\begin{table*}[ht]
\caption{Summary of the optimal hyperparameters used for the baseline and adapted models for the task of PC-CMR artefact detection. We refer the reader to Section \ref{sect:add_params}, where all hyperparameter descriptions are presented.}
  \centering
  \small
  \begin{adjustbox}{max width=\textwidth}
  \begin{tabular}{l||c|c|c|c|c|c|c|c|c|c||}
    \toprule
    \midrule
    &\multicolumn{10}{c||}{\textbf{PC-CMR}} \\
    \textbf{Method} 
     & \textbf{$epochs$} & \textbf{$l_{r}$} & \textbf{$\lambda_{ENN}$} & \textbf{$\lambda_A$} & \textbf{$\lambda_{M}$} & \textbf{$SN_{coeff}$} & \textbf{$S_{m}$} & \textbf{$\mathrm{\lambda_{E}}$} & \textbf{$\mathrm{\lambda_{D}}$} & \textbf{$\mathrm{\lambda_{u}}$}\\
        \midrule
        \textbf{1. Baseline} & 100 & $\mathrm{10^{-4}}$ & - & - & - &  & - & - &  & -\\
        \textbf{2. Baseline + AvUC} & 80 & $\mathrm{10^{-4}}$ & - & 0.6 & - &  & - & - & - & -\\
        \textbf{3. Baseline + MMCE} & 80 & $\mathrm{10^{-4}}$ & - & - & 25 &  & - & - & - & -\\
        \textbf{4. ENN} & 150 & $\mathrm{10^{-4}}$ & 40 & - & - & - & - & - & - & -\\
        \textbf{5. ENN + AvUC} & 100 & $\mathrm{10^{-4}}$ & 40 & 0.6 & - & - & - & - & - & -\\
        \textbf{6. ENN + MMCE} & 100 & $\mathrm{10^{-4}}$ & 40 & - & 40 & - & - & - & - & -\\
        \textbf{7. DDU} & 80 & $\mathrm{10^{-4}}$ & - & - & - & 0.8 & 0.1 & - & - & -\\
        \textbf{8. DDU + AvUC} & 80 & $\mathrm{10^{-4}}$ & - & 0.6 & - & 0.9 & 0.8 & - & - & -\\
        \textbf{9. DDU + MMCE} & 300 & $\mathrm{10^{-6}}$ & - & - & 50 & 0.9 & 0.8 & - & - & -\\
        \textbf{10. LDU} & 100 & $\mathrm{10^{-6}}$ & - & - & - & - & - & 0.9 & 2 & 4\\
        \textbf{11. LDU + AvUC} & 100 & $\mathrm{10^{-6}}$ & - & 0.4 & - & - & - & 0.2 & 1.2 & 5\\
        \textbf{12. LDU + MMCE} & 80 & $\mathrm{10^{-6}}$ & - & - & 50 & - & - & 0.8 & 1.5 & 5\\ 
        \bottomrule
    \end{tabular}
\end{adjustbox}
\label{table:Deter_Hyper_param_PCCMR}
\end{table*}

\begin{table*}[ht]
\caption{Summary of the optimal hyperparameters used for the baseline and adapted models for the task of cardiac disease diagnosis (ACDC dataset). We refer the reader to Section \ref{sect:add_params}, where all hyperparameter descriptions are presented.}
  \centering
  \small
  \begin{adjustbox}{max width=\textwidth}
  \begin{tabular}{l||c|c|c|c|c|c|c|c|c|c||}
    \toprule
    \midrule
    &\multicolumn{10}{c||}{\textbf{ACDC}} \\
    \textbf{Method} 
     & \textbf{$epochs$} & \textbf{$l_{r}$} & \textbf{$\lambda_{ENN}$} & \textbf{$\lambda_A$} & \textbf{$\lambda_{M}$} & \textbf{$SN_{coeff}$} & \textbf{$S_{m}$} & \textbf{$\mathrm{\lambda_{E}}$} & \textbf{$\mathrm{\lambda_{D}}$} & \textbf{$\mathrm{\lambda_{u}}$}\\
        \midrule
        \textbf{1. Baseline} & 100 & $\mathrm{10^{-3}}$ & - & - & - &  & - & - &  & -\\
        \textbf{2. Baseline + AvUC} & 60 & $\mathrm{10^{-3}}$ & - & 0.6 & - &  & - & - & - & -\\
        \textbf{3. Baseline + MMCE} & 80 & $\mathrm{10^{-3}}$ & - & - & 10 &  & - & - & - & -\\
        \textbf{4. ENN} & 60 & $\mathrm{10^{-4}}$ & 30 & - & - & - & - & - & - & -\\
        \textbf{5. ENN + AvUC} & 60 & $\mathrm{10^{-4}}$ & 35 & 2 & - & - & - & - & - & -\\
        \textbf{6. ENN + MMCE} & 60 & $\mathrm{10^{-4}}$ & 32 & - & 3 & - & - & - & - & -\\
        \textbf{7. DDU} & 150 & $\mathrm{10^{-4}}$ & - & - & - & 0.9 & 0.6 & - & - & -\\
        \textbf{8. DDU + AvUC} & 200 & $\mathrm{10^{-4}}$ & - & 0.7 & - & 0.9 & 0.8 & - & - & -\\
        \textbf{9. DDU + MMCE} & 300 & $\mathrm{10^{-4}}$ & - & - & 50 & 0.9 & 0.8 & - & - & -\\
        \textbf{10. LDU} & 80 & $\mathrm{10^{-4}}$ & - & - & - & - & - & 0.2 & 2 & 4\\
        \textbf{11. LDU + AvUC} & 80 & $\mathrm{10^{-4}}$ & - & 0.6 & - & - & - & 0.4 & 1.2 & 4\\
        \textbf{12. LDU + MMCE} & 80 & $\mathrm{10^{-3}}$ & - & - & 20 & - & - & 0.2 & 2 & 4\\ 
        \bottomrule
    \end{tabular}
\end{adjustbox}
\label{table:Deter_Hyper_param_ACDC}
\end{table*}

\subsection{Implementation Details}

All models were trained on an NVIDIA A6000 48GB GPU using an Adam optimiser. All data for both experiments were augmented with random flipping and rotations. 

\subsection{Evaluation}

We evaluated the performance of all models with a range of metrics, notably the Balanced Accuracy (BACC) for accuracy and the ECE, AECE \citep{ding2020revisiting}, and OE for calibration. For all calibration metrics, a lower score implies a better-calibrated model. 

\section{Results}
\label{sec:Results}

All accuracy and calibration measures computed are presented in Tables \ref{table:all_PCCMR} and \ref{table:all_ACDC} for the PC-CMR artefact prediction and cardiac disease diagnosis (ACDC dataset) models respectively. All best-performing metrics are indicated in bold. The experiments were run ten times and the mean and standard deviation of all metrics are reported. We also report the result of using these 10 runs as a single ensemble by averaging the prediction probabilities from each model.

\begin{table*}[ht]
\caption{Accuracy and calibration metrics for the PC-CMR artefact prediction task using the baseline model, three deterministic uncertainty models (DUM), two uncertainty-aware training approaches and their combinations. BACC: Balanced accuracy, ECE: Expected Calibration Error, AECE: Adaptive ECE, OE: Overconfidence Error.  All values are mean $\pm$ standard deviation computed over ten runs. The lowest and optimal metric across strategies is highlighted in bold. All calibration metrics should ideally move to zero as models become more calibrated as indicated by the $\Downarrow$, whereas for the BACC accuracy metric a high value is good}
\centering
  \small
  \begin{adjustbox}{max width=\textwidth}
  \begin{tabular}{l||c|c|c|c||}
    \toprule
    \midrule
    &\multicolumn{4}{c||}{\textbf{PC-CMR}}\\
    \textbf{Baseline and DUM models} 
     & \textbf{BACC \%} {$\Uparrow$} & {\textbf{ECE}} {$\Downarrow$} & \textbf{AECE} {$\Downarrow$} & \textbf{OE} {$\Downarrow$} \\
        \midrule
        \textbf{1. Baseline model} & 65 $\pm$\num{4.20} & 0.31 $\pm$\num{0.04} & 0.29 $\pm$\num{0.04} & 0.28 $\pm$\num{0.04} \\
        \textbf{2. Baseline model (Ensemble)*} & 69 & 0.20 & 0.11 & 0.13 \\
        \textbf{3. ENN} & 71 $\pm$\num{3.46} & 0.28 $\pm$\num{0.02} & 0.26 $\pm$\num{0.03} & 0.26 $\pm$\num{0.03} \\ 
        \textbf{4. DDU} & \textbf{73} $\pm$\num{1.62} & 0.23 $\pm$\num{0.03} & 0.19 $\pm$\num{0.03} & 0.19 $\pm$\num{0.03} \\ 
        \textbf{5. LDU} & 72 $\pm$\num{3.67} & \textbf{0.14} $\pm$\num{0.05} & \textbf{0.05} $\pm$\num{0.06} & \textbf{0.07} $\pm$\num{0.06}\\ 
        \bottomrule
    \textbf{AvUC uncertainty-aware training approach} 
     & \textbf{BACC \%} {$\Uparrow$} & {\textbf{ECE}}{$\Downarrow$} & \textbf{AECE} {$\Downarrow$} & \textbf{OE} {$\Downarrow$}\\
        \midrule
        \textbf{1. Baseline model + AvUC} & 69 $\pm$\num{3.35} & 0.26 $\pm$\num{0.03} & 0.23 $\pm$\num{0.05} & 0.23 $\pm$\num{0.04} \\
        \textbf{2. ENN + AvUC} & 72 $\pm$\num{2.29} & 0.24 $\pm$\num{0.08} & 0.21 $\pm$\num{0.08} & 0.22 $\pm$\num{0.07} \\ 
        \textbf{3. DDU + AvUC} & 71 $\pm$\num{2.08} & 0.20 $\pm$\num{0.02} & \textbf{0.16} $\pm$\num{0.01} & 0.16 $\pm$\num{0.02} \\ 
        \textbf{4. LDU + AvUC} & 73 $\pm$\num{3.72} & \textbf{0.11} $\pm$\num{0.03} & 0.16 $\pm$\num{0.03} & \textbf{0.02} $\pm$\num{0.01}\\ 
        \bottomrule
    \textbf{MMCE uncertainty-aware training approach} 
     & \textbf{BACC \%} {$\Uparrow$} & {\textbf{ECE}}{$\Downarrow$} & \textbf{AECE} {$\Downarrow$} & \textbf{OE} {$\Downarrow$}\\
        \midrule
        \textbf{1. Baseline model + MMCE} & 70 $\pm$\num{2.94} & 0.25 $\pm$\num{0.03} & 0.22 $\pm$\num{0.003} & 0.22 $\pm$\num{0.03} \\
        \textbf{2. ENN + MMCE} & 74 $\pm$\num{2.64} & 0.23 $\pm$\num{0.04} & 0.20 $\pm$\num{0.05} & 0.20 $\pm$\num{0.05} \\ 
        \textbf{3. DDU + MMCE} & 72 $\pm$\num{2.03} & 0.19 $\pm$\num{0.02} & 0.16 $\pm$\num{0.02} & 0.16 $\pm$\num{0.02} \\ 
        \textbf{4. LDU + MMCE} & \textbf{75} $\pm$\num{2.00} & \textbf{0.11} $\pm$\num{0.02} & \textbf{0.05} $\pm$\num{0.04} & \textbf{0.02} $\pm$\num{0.03}\\ 
        \bottomrule
    \end{tabular}
    \end{adjustbox}
    \begin{threeparttable}
    \begin{tablenotes}
    \centering
         \item $^*$ \textit{Ensemble over the 10 baseline model runs.}
    \end{tablenotes} 
    \end{threeparttable}
\label{table:all_PCCMR}
\end{table*}

\begin{table*}[ht]
\caption{Accuracy and calibration metrics for the ACDC diagnosis task using the baseline model, three deterministic uncertainty models (DUM) two uncertainty-aware approaches, and their combinations. BACC: Balanced Accuracy, ECE: Expected Calibration Error, AECE: Adaptive ECE, OE: Overconfidence Error.  All values are mean $\pm$ standard deviation computed over ten runs. The lowest and optimal metric across strategies is highlighted in bold. All calibration metrics should ideally move to zero as models become more calibrated as indicated by the $\Downarrow$, whereas for the BACC accuracy metric a high value is good.}
\centering
  \small
  \begin{adjustbox}{max width=\textwidth}
  \begin{tabular}{l||c|c|c|c||}
    \toprule
    \midrule
    &\multicolumn{4}{c||}{\textbf{ACDC}}\\
    \textbf{Baseline and DUM models} 
     & \textbf{BACC \%} {$\Uparrow$} & {\textbf{ECE}} {$\Downarrow$} & \textbf{AECE} {$\Downarrow$} & \textbf{OE}{$\Downarrow$}\\
        \midrule
        \textbf{1. Baseline model} & 74 $\pm$\num{1.24} & 0.18 $\pm$\num{0.02} & 0.15 $\pm$\num{0.02} & 0.08 $\pm$\num{0.02} \\
        \textbf{2. Baseline model (Ensemble)*} & 80 & 0.15 & 0.13 & 0.01 \\
        \textbf{3. Evidential Neural Network} & 74 $\pm$\num{4.28} & 0.16 $\pm$\num{0.02} & 0.15 $\pm$\num{0.02} & 0.14 $\pm$\num{0.02} \\
        \textbf{4. Deep Deterministic Framework} & \textbf{76} $\pm$\num{3.03} & 0.13 $\pm$\num{0.03} & 0.10 $\pm$\num{0.03} & 0.07 $\pm$\num{0.02} \\
        \textbf{5. Latent Discriminant Uncertainty Framework}& 75 $\pm$\num{2.19} & \textbf{0.12} $\pm$\num{0.02} & \textbf{0.09} $\pm$\num{0.04} & \textbf{0.05} $\pm$\num{0.02} \\
        \bottomrule
    \textbf{AvUC uncertainty-aware approaches} 
     & \textbf{BACC \%} {$\Uparrow$} & {\textbf{ECE}}{$\Downarrow$} & \textbf{AECE} {$\Downarrow$} & \textbf{OE} {$\Downarrow$}\\
        \midrule
        \textbf{1. Baseline model + AvUC} & \textbf{81} $\pm$\num{3.82} & 0.14 $\pm$\num{0.02} & 0.12 $\pm$\num{0.02} & 0.11 $\pm$\num{0.02} \\
        \textbf{2. Evidential Neural Network + AvUC} & 75 $\pm$\num{4.15} & 0.17 $\pm$\num{0.02} & 0.16 $\pm$\num{0.02} & 0.14 $\pm$\num{0.01} \\ 
        \textbf{3. Deep Deterministic Uncertainty + AvUC} & 73 $\pm$\num{1.89} & \textbf{0.12} $\pm$\num{0.01} & \textbf{0.09} $\pm$\num{0.01} & \textbf{0.06} $\pm$\num{0.01} \\ 
        \textbf{4. Latent Discriminant Uncertainty + AvUC} & 74 $\pm$\num{2.72} & 0.14 $\pm$\num{0.02} & 0.11 $\pm$\num{0.02} & 0.07 $\pm$\num{0.02}\\ 
        \bottomrule
    \textbf{MMCE uncertainty-aware approaches} 
     & \textbf{BACC \%} {$\Uparrow$} & {\textbf{ECE}}{$\Downarrow$} & \textbf{AECE} {$\Downarrow$} & \textbf{OE} {$\Downarrow$}\\
        \midrule
        \textbf{1. Baseline model + MMCE} & 80 $\pm$\num{2.10} & 0.13 $\pm$\num{0.01} & 0.12 $\pm$\num{0.01} & 0.10 $\pm$\num{0.01} \\
        \textbf{2. Evidential Neural Network + MMCE} & 78 $\pm$\num{2.26} & 0.16 $\pm$\num{0.01} & 0.15 $\pm$\num{0.02} & 0.13 $\pm$\num{0.01} \\ 
        \textbf{3. Deep Deterministic Uncertainty + MMCE} & 76 $\pm$\num{1.91} & \textbf{0.12} $\pm$\num{0.01} & \textbf{0.08} $\pm$\num{0.02} & \textbf{0.07} $\pm$\num{0.01} \\ 
        \textbf{4. Latent Discriminant Uncertainty + MMCE} & \textbf{81} $\pm$\num{2.50} & 0.13 $\pm$\num{0.02} & 0.10 $\pm$\num{0.02} & 0.09 $\pm$\num{0.02}\\ 
        \bottomrule
    \end{tabular}
    \end{adjustbox}
    \begin{threeparttable}
    \begin{tablenotes}
    \centering
         \item $^*$ \textit{Ensemble over the 10 baseline model runs.}
    \end{tablenotes} 
    \end{threeparttable}
\label{table:all_ACDC}
\end{table*} 

\section{Discussion}
\label{sec:Discussion}

In this paper, we have evaluated three different DUM approaches and two uncertainty-aware training approaches against a standard baseline ResNet50 architecture and also investigated the utility of combining these approaches into a novel deterministic uncertainty-aware training approach. All approaches were evaluated with two clinical CMR-based cardiac applications. 

\subsection{Model Performance}

Our results indicate that for both the PC-CMR and ACDC experiments the DUM and uncertainty-aware training approaches improved calibration and sometimes accuracy. Overall, the DUMs achieved slightly better improvements to calibration than uncertainty-aware training, although performance varied from technique to technique.
In particular, the DDU and LDU models performed best for both accuracy and calibration performance. The classical approach of model ensembling also improved outcomes for both tasks.  When DUMs and uncertainty-aware training were combined, some further improvements were achieved in accuracy and/or calibration. However, this was not the case for all DUMs, notably the LDU model. This may be attributed to the inherent inclusion of uncertainty-based losses already mathematically catered for within the design of these networks. Overall though, combining DUMs with uncertainty-aware training strategies did result in some improvements in calibration performance, with the MMCE loss fairing slightly better in this respect. Noticeably, the ENN had a poorer performance with respect to calibration performance for the ACDC dataset, which may imply the network is not optimal for multi-class classification problems. 

As we have reported in our previous work, \cite{dawood2023uncertainty}, there is still no consensus on which metric best quantifies calibration performance. However, our analysis indicates that the DUMs' architectural changes enhance the models' abilities to quantify uncertainty and improve model calibration. Furthermore, our findings suggest that the MMCE loss as an uncertainty-aware training strategy improves calibration with no negative impact on accuracy. This may indicate that a model which can be viewed as a differentiable proxy to ECE provides a more calibration-aware network and can in the absence of a DUM improve both accuracy and calibration of DL models for cardiac AI applications. \\

\subsection{Limitations and Future Work}

Calibration performance is known to suffer under domain shifts in the presence of OOD data \citep{karimi2022improving}. In this paper, we have focused only on evaluation within datasets (i.e. internal validation). Therefore, future work should focus on repeating our analysis in scenarios with domain shifts and/or OOD data. It would also be useful to disentangle and evaluate the epistemic and aleatoric uncertainties, which we have not investigated in this paper. Future work should also investigate other medical imaging applications in which DL models are likely to be used in a decision-support setting.

\section{Conclusion}

In summary, we have investigated different DUMs with and without uncertainty-aware training strategies and evaluated all models with accuracy and calibration metrics. Overall our analysis indicates that a DUM network can improve both accuracy and calibration performance, with improved outcomes when compared to the baseline model and the standard ensemble approach. We hope that the work we have presented can contribute to furthering the clinical translation of trustworthy DL models in medical imaging applications.   



\acks{This work was supported by the Kings DRIVE Health CDT for Data-Driven Health and further funded/supported by the National Institute for Health Research (NIHR) Biomedical Research Centre at Guy’s and St Thomas’ NHS Foundation Trust and King’s College London. Additionally, this research was funded in whole, or in part, by the Wellcome Trust WT203148/Z/16/Z. For the purpose of open access, the author has applied a CC BY public copyright license to any author-accepted manuscript version arising from this submission. The work was also supported by the EPSRC through the SmartHeart Programme Grant (EP/P001009/1). This research has been conducted using the UK Biobank Resource under Application Number 17806. 

The views expressed in this paper are those of the authors and not necessarily those of the NHS, EPSRC, the NIHR, or the Department of Health and Social Care.}

\ethics{The work follows appropriate ethical standards in conducting research and writing the manuscript, following all applicable laws and regulations regarding the treatment of animals or human subjects.}

This research study for the PC-CMR data was conducted retrospectively using human subject data made available using the UK Biobank Resource under Application Number 17806. All relevant protocols were adhered to in order to retrieve and store the patient data and related images.

Ethical approval for the use of the GSTFT dataset was obtained from the London Dulwich Research Ethics Committee (ID: 19/LO/1957). The other datasets are available online, under previous approval from the various host institute ethical committees.


\coi{We declare we don't have conflicts of interest.}

\bibliography{Uncertainty/refs}

\begin{thebibliography}{38}
\providecommand{\natexlab}[1]{#1}
\providecommand{\url}[1]{\texttt{#1}}
\expandafter\ifx\csname urlstyle\endcsname\relax
  \providecommand{\doi}[1]{doi: #1}\else
  \providecommand{\doi}{doi: \begingroup \urlstyle{rm}\Url}\fi

\bibitem[Abdar et~al.(2021)Abdar, Pourpanah, Hussain, et~al.]{abdar2021review}
Moloud Abdar, Farhad Pourpanah, Sadiq Hussain, et~al.
\newblock A review of uncertainty quantification in deep learning: Techniques, applications and challenges.
\newblock \emph{Information Fusion}, 2021.

\bibitem[Bernard et~al.(2018)Bernard, Lalande, Zotti, Cervenansky, Yang, Heng, Cetin, Lekadir, Camara, Ballester, et~al.]{bernard2018deep}
Olivier Bernard, Alain Lalande, Clement Zotti, Frederick Cervenansky, Xin Yang, Pheng-Ann Heng, Irem Cetin, Karim Lekadir, Oscar Camara, Miguel Angel~Gonzalez Ballester, et~al.
\newblock Deep learning techniques for automatic mri cardiac multi-structures segmentation and diagnosis: is the problem solved?
\newblock \emph{IEEE transactions on medical imaging}, 37\penalty0 (11):\penalty0 2514--2525, 2018.

\bibitem[Buddenkotte et~al.(2023)Buddenkotte, Sanchez, Crispin-Ortuzar, Woitek, McCague, Brenton, {\"O}ktem, Sala, and Rundo]{buddenkotte2023calibrating}
Thomas Buddenkotte, Lorena~Escudero Sanchez, Mireia Crispin-Ortuzar, Ramona Woitek, Cathal McCague, James~D Brenton, Ozan {\"O}ktem, Evis Sala, and Leonardo Rundo.
\newblock Calibrating ensembles for scalable uncertainty quantification in deep learning-based medical image segmentation.
\newblock \emph{Computers in Biology and Medicine}, page 107096, 2023.

\bibitem[Cao et~al.(2021)Cao, Chen, Li, Peng, Wang, and Cheng]{9214845}
Xuyang Cao, Houjin Chen, Yanfeng Li, Yahui Peng, Shu Wang, and Lin Cheng.
\newblock Uncertainty aware temporal-ensembling model for semi-supervised abus mass segmentation.
\newblock \emph{IEEE Transactions on Medical Imaging}, 40\penalty0 (1):\penalty0 431--443, 2021.
\newblock \doi{10.1109/TMI.2020.3029161}.

\bibitem[Chan et~al.(2022)Chan, O’Hanlon, Marquez, Petalcorin, Mariscal-Harana, Gu, Kim, Judd, Chowienczyk, Schnabel, et~al.]{chan2022automated}
Emily Chan, Ciaran O’Hanlon, Carlota~Asegurado Marquez, Marwenie Petalcorin, Jorge Mariscal-Harana, Haotian Gu, Raymond~J Kim, Robert~M Judd, Phil Chowienczyk, Julia~A Schnabel, et~al.
\newblock Automated quality controlled analysis of 2d phase contrast cardiovascular magnetic resonance imaging.
\newblock In \emph{International Workshop on Statistical Atlases and Computational Models of the Heart}, pages 101--111. Springer, 2022.

\bibitem[Dawood et~al.(2023{\natexlab{a}})Dawood, Chan, Razavi, King, and Puyol-Anton]{dawood2023addressing}
Tareen Dawood, Emily Chan, Reza Razavi, Andrew~P King, and Esther Puyol-Anton.
\newblock Addressing deep learning model calibration using evidential neural networks and uncertainty-aware training.
\newblock \emph{arXiv preprint arXiv:2301.13296}, 2023{\natexlab{a}}.

\bibitem[Dawood et~al.(2023{\natexlab{b}})Dawood, Chen, Sidhu, Ruijsink, Gould, Porter, Elliott, Mehta, Rinaldi, Puyol-Ant{\'o}n, et~al.]{dawood2023uncertainty}
Tareen Dawood, Chen Chen, Baldeep~S Sidhu, Bram Ruijsink, Justin Gould, Bradley Porter, Mark~K Elliott, Vishal Mehta, Christopher~A Rinaldi, Esther Puyol-Ant{\'o}n, et~al.
\newblock Uncertainty aware training to improve deep learning model calibration for classification of cardiac mr images.
\newblock \emph{Medical Image Analysis}, page 102861, 2023{\natexlab{b}}.

\bibitem[Ding et~al.(2020)Ding, Liu, Xiong, and Shi]{ding2020revisiting}
Yukun Ding, Jinglan Liu, Jinjun Xiong, and Yiyu Shi.
\newblock Revisiting the evaluation of uncertainty estimation and its application to explore model complexity-uncertainty trade-off.
\newblock In \emph{Proceedings of the IEEE/CVF Conference on Computer Vision and Pattern Recognition Workshops}, pages 4--5, 2020.

\bibitem[Franchi et~al.(2022)Franchi, Yu, Bursuc, Aldea, Dubuisson, and Filliat]{franchi2022latent}
Gianni Franchi, Xuanlong Yu, Andrei Bursuc, Emanuel Aldea, Severine Dubuisson, and David Filliat.
\newblock Latent discriminant deterministic uncertainty.
\newblock In \emph{European Conference on Computer Vision}, pages 243--260. Springer, 2022.

\bibitem[Gawlikowski et~al.(2021)Gawlikowski, Tassi, Ali, Lee, Humt, Feng, Kruspe, Triebel, Jung, Roscher, et~al.]{gawlikowski2021survey}
Jakob Gawlikowski, Cedrique Rovile~Njieutcheu Tassi, Mohsin Ali, Jongseok Lee, Matthias Humt, Jianxiang Feng, Anna Kruspe, Rudolph Triebel, Peter Jung, Ribana Roscher, et~al.
\newblock A survey of uncertainty in deep neural networks.
\newblock \emph{arXiv preprint arXiv:2107.03342}, 2021.

\bibitem[Guo et~al.(2017)Guo, Pleiss, Sun, and Weinberger]{guo2017calibration}
Chuan Guo, Geoff Pleiss, Yu~Sun, and Kilian~Q Weinberger.
\newblock On calibration of modern neural networks.
\newblock In \emph{International Conference on Machine Learning}, pages 1321--1330. PMLR, 2017.

\bibitem[Karandikar et~al.(2021)Karandikar, Cain, Tran, Lakshminarayanan, Shlens, Mozer, and Roelofs]{karandikar2021soft}
Archit Karandikar, Nicholas Cain, Dustin Tran, Balaji Lakshminarayanan, Jonathon Shlens, Michael~C Mozer, and Rebecca Roelofs.
\newblock Soft calibration objectives for neural networks.
\newblock \emph{Advances in Neural Information Processing Systems}, 34, 2021.

\bibitem[Karimi and Gholipour(2022)]{karimi2022improving}
Davood Karimi and Ali Gholipour.
\newblock Improving calibration and out-of-distribution detection in deep models for medical image segmentation.
\newblock \emph{IEEE Transactions on Artificial Intelligence}, 4\penalty0 (2):\penalty0 383--397, 2022.

\bibitem[Krishnan and Tickoo(2020)]{krishnan2020improving}
Ranganath Krishnan and Omesh Tickoo.
\newblock Improving model calibration with accuracy versus uncertainty optimization.
\newblock \emph{Advances in Neural Information Processing Systems}, 2020.

\bibitem[Kumar and Sarawagi(2019)]{kumar2019calibration}
Aviral Kumar and Sunita Sarawagi.
\newblock Calibration of encoder decoder models for neural machine translation.
\newblock \emph{arXiv preprint arXiv:1903.00802}, 2019.

\bibitem[Kumar et~al.(2018)Kumar, Sarawagi, and Jain]{kumar2018trainable}
Aviral Kumar, Sunita Sarawagi, and Ujjwal Jain.
\newblock Trainable calibration measures for neural networks from kernel mean embeddings.
\newblock In \emph{International Conference on Machine Learning}, pages 2805--2814. PMLR, 2018.

\bibitem[Lakshminarayanan et~al.(2017)Lakshminarayanan, Pritzel, and Blundell]{lakshminarayanan2017simple}
Balaji Lakshminarayanan, Alexander Pritzel, and Charles Blundell.
\newblock Simple and scalable predictive uncertainty estimation using deep ensembles.
\newblock \emph{Advances in neural information processing systems}, 30, 2017.

\bibitem[Maier-Hein et~al.(2023)Maier-Hein, Reinke, Godau, Tizabi, B{\"u}ttner, Christodoulou, Glocker, Isensee, Kleesiek, Kozubek, et~al.]{maier2206metrics}
L~Maier-Hein, A~Reinke, P~Godau, M~Tizabi, F~B{\"u}ttner, E~Christodoulou, B~Glocker, F~Isensee, J~Kleesiek, M~Kozubek, et~al.
\newblock Metrics reloaded: Recommendations for image analysis validation. arxiv 2023.
\newblock \emph{arXiv preprint arXiv:2206.01653}, 2023.

\bibitem[Malinin and Gales(2018)]{malinin2018predictive}
Andrey Malinin and Mark Gales.
\newblock Predictive uncertainty estimation via prior networks.
\newblock \emph{arXiv preprint arXiv:1802.10501}, 2018.

\bibitem[Miyato et~al.(2018)Miyato, Kataoka, Koyama, and Yoshida]{miyato2018spectral}
Takeru Miyato, Toshiki Kataoka, Masanori Koyama, and Yuichi Yoshida.
\newblock Spectral normalization for generative adversarial networks.
\newblock \emph{arXiv preprint arXiv:1802.05957}, 2018.

\bibitem[Mukhoti et~al.(2020)Mukhoti, Kulharia, Sanyal, Golodetz, Torr, and Dokania]{mukhoti2020calibrating}
Jishnu Mukhoti, Viveka Kulharia, Amartya Sanyal, Stuart Golodetz, Philip Torr, and Puneet Dokania.
\newblock Calibrating deep neural networks using focal loss.
\newblock \emph{Advances in Neural Information Processing Systems}, 33:\penalty0 15288--15299, 2020.

\bibitem[Mukhoti et~al.(2021{\natexlab{a}})Mukhoti, Kirsch, van Amersfoort, Torr, and Gal]{mukhoti2021deep}
Jishnu Mukhoti, Andreas Kirsch, Joost van Amersfoort, Philip~HS Torr, and Yarin Gal.
\newblock Deep deterministic uncertainty: A simple baseline.
\newblock \emph{arXiv preprint arXiv:2102.11582}, 2021{\natexlab{a}}.

\bibitem[Mukhoti et~al.(2021{\natexlab{b}})Mukhoti, Kirsch, van Amersfoort, Torr, and Gal]{mukhoti2021deterministic}
Jishnu Mukhoti, Andreas Kirsch, Joost van Amersfoort, Philip~HS Torr, and Yarin Gal.
\newblock Deterministic neural networks with inductive biases capture epistemic and aleatoric uncertainty.
\newblock \emph{arXiv preprint arXiv:2102.11582}, 2, 2021{\natexlab{b}}.

\bibitem[Platt et~al.(1999)]{platt1999probabilistic}
John Platt et~al.
\newblock Probabilistic outputs for support vector machines and comparisons to regularized likelihood methods.
\newblock \emph{Advances in large margin classifiers}, 10\penalty0 (3):\penalty0 61--74, 1999.

\bibitem[Postels et~al.(2021)Postels, Segu, Sun, Sieber, Van~Gool, Yu, and Tombari]{postels2021practicality}
Janis Postels, Mattia Segu, Tao Sun, Luca Sieber, Luc Van~Gool, Fisher Yu, and Federico Tombari.
\newblock On the practicality of deterministic epistemic uncertainty.
\newblock \emph{arXiv preprint arXiv:2107.00649}, 2021.

\bibitem[Rahaman et~al.(2021)]{rahaman2021uncertainty}
Rahul Rahaman et~al.
\newblock Uncertainty quantification and deep ensembles.
\newblock \emph{Advances in Neural Information Processing Systems}, 34:\penalty0 20063--20075, 2021.

\bibitem[Roelofs et~al.(2022)Roelofs, Cain, Shlens, and Mozer]{roelofs2022mitigating}
Rebecca Roelofs, Nicholas Cain, Jonathon Shlens, and Michael~C Mozer.
\newblock Mitigating bias in calibration error estimation.
\newblock In \emph{International Conference on Artificial Intelligence and Statistics}, pages 4036--4054. PMLR, 2022.

\bibitem[Sensoy et~al.(2018)Sensoy, Kaplan, and Kandemir]{sensoy2018evidential}
Murat Sensoy, Lance Kaplan, and Melih Kandemir.
\newblock Evidential deep learning to quantify classification uncertainty.
\newblock \emph{arXiv preprint arXiv:1806.01768}, 2018.

\bibitem[Sensoy et~al.(2020)Sensoy, Kaplan, Cerutti, and Saleki]{sensoy2020uncertainty}
Murat Sensoy, Lance Kaplan, Federico Cerutti, and Maryam Saleki.
\newblock Uncertainty-aware deep classifiers using generative models.
\newblock In \emph{Proceedings of the AAAI Conference on Artificial Intelligence}, volume~34, pages 5620--5627, 2020.

\bibitem[Sensoy et~al.(2021)Sensoy, Saleki, Julier, Aydogan, and Reid]{sensoy2021misclassification}
Murat Sensoy, Maryam Saleki, Simon Julier, Reyhan Aydogan, and John Reid.
\newblock Misclassification risk and uncertainty quantification in deep classifiers.
\newblock In \emph{Proceedings of the IEEE/CVF Winter Conference on Applications of Computer Vision}, pages 2484--2492, 2021.

\bibitem[Sudlow et~al.(2015)Sudlow, Gallacher, Allen, Beral, Burton, Danesh, Downey, Elliott, Green, Landray, et~al.]{sudlow2015uk}
Cathie Sudlow, John Gallacher, Naomi Allen, Valerie Beral, Paul Burton, John Danesh, Paul Downey, Paul Elliott, Jane Green, Martin Landray, et~al.
\newblock Uk biobank: an open access resource for identifying the causes of a wide range of complex diseases of middle and old age.
\newblock \emph{PLoS medicine}, 12\penalty0 (3):\penalty0 e1001779, 2015.

\bibitem[Thiagarajan et~al.(2022)Thiagarajan, Thopalli, Rajan, and Turaga]{thiagarajan2022training}
Jayaraman~J Thiagarajan, Kowshik Thopalli, Deepta Rajan, and Pavan Turaga.
\newblock Training calibration-based counterfactual explainers for deep learning models in medical image analysis.
\newblock \emph{Scientific reports}, 12\penalty0 (1):\penalty0 597, 2022.

\bibitem[Thulasidasan et~al.(2019)Thulasidasan, Chennupati, Bilmes, Bhattacharya, and Michalak]{thulasidasan2019mixup}
Sunil Thulasidasan, Gopinath Chennupati, Jeff Bilmes, Tanmoy Bhattacharya, and Sarah Michalak.
\newblock On mixup training: Improved calibration and predictive uncertainty for deep neural networks.
\newblock \emph{arXiv preprint arXiv:1905.11001}, 2019.

\bibitem[Wang(2023)]{wang2023calibration}
Cheng Wang.
\newblock Calibration in deep learning: A survey of the state-of-the-art.
\newblock \emph{arXiv preprint arXiv:2308.01222}, 2023.

\bibitem[Wibowo et~al.(2022)Wibowo, Triadyaksa, Sugiharto, Sarwoko, Nugroho, Arai, and Kawakubo]{wibowo2022cardiac}
Adi Wibowo, Pandji Triadyaksa, Aris Sugiharto, Eko~Adi Sarwoko, Fajar~Agung Nugroho, Hideo Arai, and Masateru Kawakubo.
\newblock Cardiac disease classification using two-dimensional thickness and few-shot learning based on magnetic resonance imaging image segmentation.
\newblock \emph{Journal of Imaging}, 8\penalty0 (7):\penalty0 194, 2022.

\bibitem[Yu et~al.(2019)Yu, Wang, Li, Fu, and Heng]{yu2019uncertainty}
Lequan Yu, Shujun Wang, Xiaomeng Li, Chi-Wing Fu, and Pheng-Ann Heng.
\newblock Uncertainty-aware self-ensembling model for semi-supervised {3D} left atrium segmentation.
\newblock In \emph{International Conference on Medical Image Computing and Computer-Assisted Intervention}, pages 605--613. Springer, 2019.

\bibitem[Zhang et~al.(2020)Zhang, Kailkhura, and Han]{zhang2020mix}
Jize Zhang, Bhavya Kailkhura, and T~Yong-Jin Han.
\newblock Mix-n-match: Ensemble and compositional methods for uncertainty calibration in deep learning.
\newblock In \emph{International conference on machine learning}, pages 11117--11128. PMLR, 2020.

\bibitem[Zou et~al.(2023)Zou, Chen, Yuan, Shen, Wang, and Fu]{zou2023review}
Ke~Zou, Zhihao Chen, Xuedong Yuan, Xiaojing Shen, Meng Wang, and Huazhu Fu.
\newblock A review of uncertainty estimation and its application in medical imaging.
\newblock \emph{arXiv preprint arXiv:2302.08119}, 2023.

\end{thebibliography}


\end{document}